\documentclass{article}
\usepackage{spconf,amsmath,graphicx}
\usepackage{algorithm}
\usepackage{algpseudocode}
\usepackage{multirow}
\usepackage{stfloats}
\usepackage{booktabs}
\usepackage{siunitx}
\usepackage{tabularx}
\usepackage[utf8]{inputenc}
\usepackage{subcaption}
\usepackage{siunitx}   % For aligning numbers by decimal point
\usepackage[hidelinks]{hyperref}
\usepackage{threeparttable} % For adding notes below the table
\title{P2DT: Mitigating Forgetting in task-incremental Learning with progressive prompt Decision Transformer}

\name{Zhiyuan Wang$^{1,2\dag}$, Xiaoyang Qu$^{2*}$, Jing Xiao$^{2}$,  Bokui Chen$^{1,3*}$, Jianzong Wang$^{2}$
\thanks{$\dag$ Work done as an intern at Ping An Technology (Shenzhen) Co., Ltd }
\thanks{$*$ Corresponding authors: Xiaoyang Qu (quxiaoy@gmail.com), Bokui Chen (Chen.bokui@sz.tsinghua.edu.cn)}
}
\address{$^{1}$Tsinghua Shenzhen International Graduate School, Tsinghua University, China \\
$^{2}$Ping An Technology (Shenzhen) Co., Ltd., Shenzhen, China \\
$^{3}$Peng Cheng Laboratory, Shenzhen, China}

\begin{document}

\maketitle
\begin{abstract}
Catastrophic forgetting poses a substantial challenge for managing intelligent agents controlled by 
a large model, causing performance degradation when these agents face new tasks. In our work, we propose a novel solution - the \textbf{P}rogressive \textbf{P}rompt \textbf{D}ecision \textbf{T}ransformer (P2DT). This method enhances a transformer-based model by dynamically appending decision tokens during new task training, thus fostering task-specific policies. Our approach mitigates forgetting in continual and offline reinforcement learning scenarios. Moreover, P2DT leverages trajectories collected via traditional reinforcement learning from all tasks and generates new task-specific tokens during training, thereby retaining knowledge from previous studies. Preliminary results demonstrate that our model effectively alleviates catastrophic forgetting and scales well with increasing task environments.Our implementation is currently available at \href{https://github.com/XiaoAI1989/P2DT}{https://github.com/XiaoAI1989/P2DT}.

\end{abstract}
\section{Introduction}
\label{sec:intro}

Artificial intelligence is critical in addressing complex control tasks in various fields, including robotics and autonomous driving. Previous studies \cite{chen2021decision,lee2022multi,zheng2022online} have applied Transformer architectures for decision-making, exploiting these architectures to process time-series models derived from offline datasets. Nevertheless, using these systems in dynamic environments, especially those requiring continual learning, is impeded by a significant obstacle termed `catastrophic forgetting,' resulting in diminished proficiency in earlier learned tasks \cite{hadsell2020embracing,parisi2019continual}. This limitation becomes particularly noticeable in multi-task learning environments. Here, several agents with unique state variables and action spaces must constantly adapt and learn. Therefore, we propose the Progressive Prompt Decision Transformer, a novel approach designed to counteract the issue of catastrophic forgetting, thus improving the performance of AI systems in continual learning environments.

Efforts to address catastrophic forgetting have resulted in the emergence of various continual learning strategies. These strategies are generally classified into two groups: regularization-oriented and rehearsal-oriented. Regularization-oriented strategies \cite{li2017learning,kirkpatrick2017overcoming,liu2023fedet} focus on restricting the modification of parameters when acquiring new skills. In contrast, this is achieved by integrating supplementary terms into the loss function, penalizing modifications potentially detrimental to prior task performance. Rehearsal-oriented strategies \cite{lopez2017gradient,rebuffi2017icarl,wang2023shoggoth} combat the issue of catastrophic forgetting by keeping a selection of data from former tasks and blending this data with novel task information throughout the training phase, which mirror the way humans continually practice to maintain knowledge over time.

Prompt learning strategically modifies input text to provide language models with insights specific to each task. Designing effective hand-crafted prompts can be a complex process that often requires heuristic approaches. Methods like soft prompting \cite{lester2021power} and prefix tuning \cite{li2021prefix} address this issue by introducing learnable parameters that guide the model's behavior, simplifying the adaptation of large language models to specific tasks. 

We enhance the Transformer-based decision model to overcome these hurdles by dynamically appending decision tokens during new task training, thus fostering task-specific policies. The main contributions are shown as follows:
\begin{itemize}
\item We propose the Progressive Decision Transformer, an innovative method that reduces catastrophic forgetting in multi-task scenarios. 
% This technique dynamically adds decision tokens, facilitating task-specific policy learning.
\item Our method optimizes by utilizing trajectories collected from all tasks for improved policy learning and generating new task-specific tokens while learning new tasks.
\item Through empirical tests, we demonstrate that our approach maintains a high level of performance with minimal reduction.
\end{itemize}

\section{The Proposed Method}
\label{sec:system}

\begin{figure*}[htb]
    \centering
    % 左侧部分
    \includegraphics[width=\linewidth,height=0.4878\linewidth]{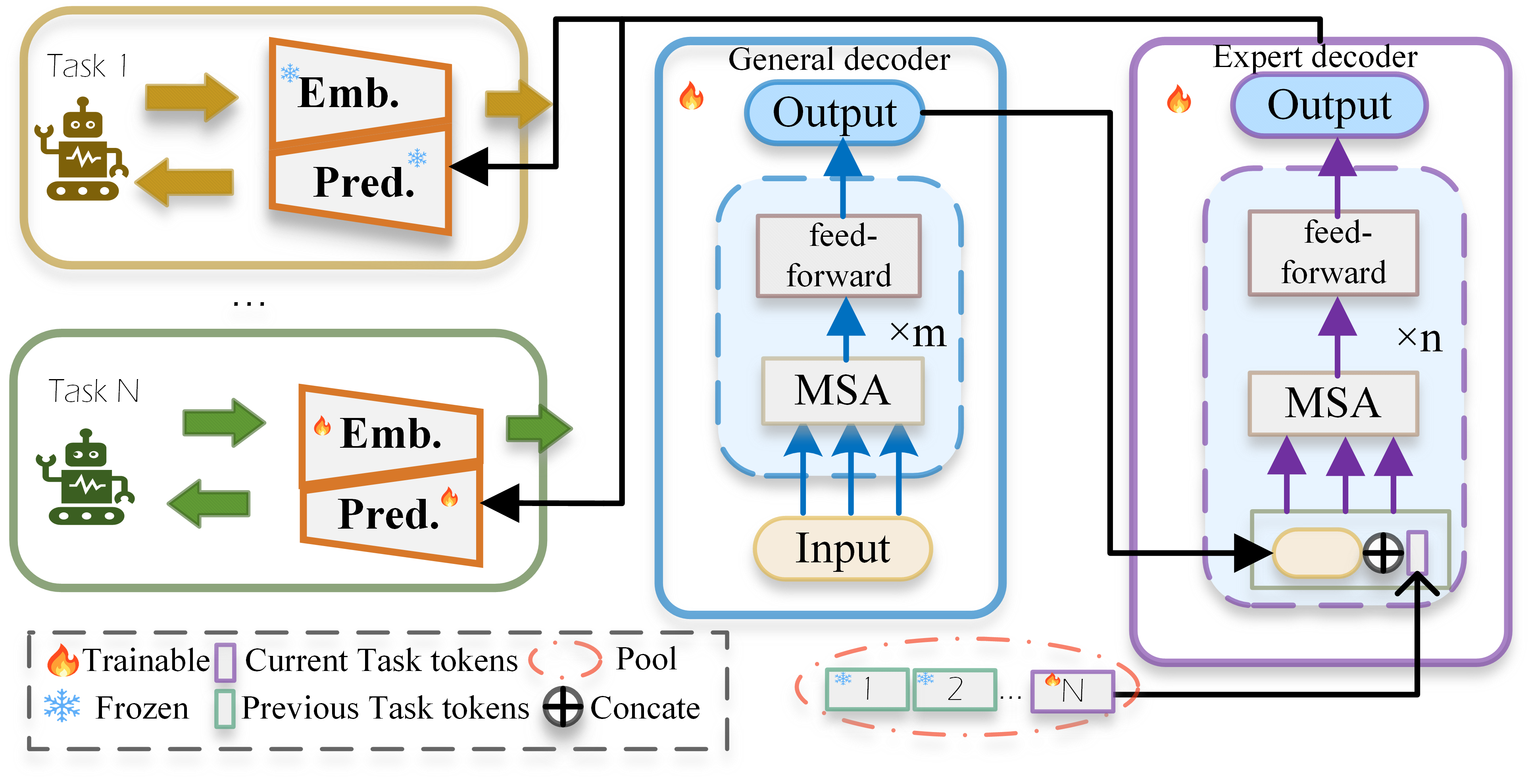}
    \caption{The framework of our algorithm. A trajectory is embedded via a linear projection. The resulting tokens from these segments are processed by m consecutive General Attention Blocks (GAB). The processed tokens are then presented to the Expert Attention Blocks (EAB) for each task.}
    \label{fig:p2dt}

    \label{fig:combined}
\end{figure*}

\subsection{Problem Formulation }
\label{ssec:prob}

In incremental decision tasks, catastrophic forgetting emerges as a crucial challenge, particularly highlighted in the scenario where an agent learns multiple tasks sequentially. This phenomenon underscores the agent's struggle to retain knowledge from earlier tasks while acquiring new information, highlighting the core problem of catastrophic forgetting in sequential task learning environments.

We extend the traditional Markov Decision Process (MDP) framework to accommodate a sequence of MDPs. The problem is defined by a sequence of tuples $(S^i, A^i, P^i, R^i)_{i=1}^N$, where $i$ indexes the task and $N$ is the number of tasks. Here, $S^i$ denotes the state space, $A^i$ represents the action space, $P^i$ signifies the transition problities, defined as $P^i(s_t^{'i}|s_t^i, a_t^i)$, and $R^i$ stands for the reward function for task $i$, denoted as $r_t^i = R^i(s_t^i, a_t^i)$. We use $s_t^i$, $a_t^i$, and $r_t^i = R(s_t^i; a_t^i)$ to denote the state, action, and reward at timestep $t$. In a continual learning framework, the objective is for the agent to learn policies $\pi^i$ that maximizes the expected return for each task, where the expected return for task $i$ is defined as $E[\sum_{t=1}^{T} \gamma_{t-1}r_{t}^i]$ with $\gamma$ being the discount factor.

This leads to a trajectory representation that is well-suited for the autoregressive training:
\begin{equation}
\mathcal{T}^i = (\hat{R}_1^i, s_1^i, a_1^i, \hat{R}_2^i, s_2^i, a_2^i, . . . , \hat{R}_T^i , s_T^i , a_T^i)
\end{equation}
This trajectory representation allows the model to understand and learn the complex dynamics of multi-agent control tasks in an offline reinforcement learning setting.

\subsection{Model Architecture }
\label{ssec:arch}

As illustrated in Fig.~\ref{fig:p2dt}, We propose the Progressive Prompt Decision Transformer (P2DT), which introduces a unique mechanism to handle reinforcement learning tasks in a continual learning setting. It draws inspiration from Complementary Learning theory \cite{kumaran2016learning} and progressive prompts in language models \cite{razdaibiedina2022progressive}. In P2DT, the general block serves as a neural analog of the neocortex, acquiring task-invariant knowledge shared across tasks. Conversely, the expert block, equipped with task-specific prompts, mimics the function of the hippocampus, acquiring task-specific knowledge. These task prompts aim to improve the model's understanding of each task while retaining knowledge from previous tasks.

Each task is associated with unique task prompts $p_i$, learned during the training phase. As new tasks are encountered, these task prompts are dynamically added to the network and placed at the end of the input sequence for the transformer's expert block. The general block of the P2DT shares an identical structure with the original Decision Transformer. This block is shared across all tasks and is responsible for learning the general patterns present in the data. 
Conversely, the expert block, which is unique for each task, is responsible for acquiring task-specific knowledge. 

The task prompt $p_i$ for task $i$ is a learned embedding included in the expert block's input. As new tasks are introduced, new task prompts are dynamically added to the model, allowing it to continually expand its capacity for learning new tasks. This is done without significantly increasing the complexity of the model, as the task prompts exhibit considerably less magnitude than the model parameters.

The output of the P2DT is computed by passing the result of the general block and the task prompt through the expert block:

\begin{equation}
\text{F}(s_{1:t}, a_{1:t}, r_{1:t}, p_i) = \text{E}(\text{G}(s_{1:t}, a_{1:t}, r_{1:t}), p_i)
\end{equation}
where $\text{G}$ denotes the function for general knowledge acquisition across tasks, $\text{E}$ corresponds to the expert function for task-specific processing, and $\text{F}$ represents the final integrated output of the system.

\subsection{Loss Function}

The P2DT is trained using a modified version of the loss function used in the Decision Transformer. The goal is to minimize the disparity between the predicted rewards and the real rewards, given the states, actions, and task prompts.
The loss function for task $i$ is given by:

\begin{equation}
\mathcal{L}_i =  \left| F(s_{1:t}, a_{1:t}, r_{1:t}, p_i) - r_{t+1:T} \right|
\end{equation}
where $(s, a, r)$ are the state, action, and reward sequences, and $p_i$ is the task prompt for task $i$.

Inspired by \cite{kirkpatrick2017overcoming}, we add a term to the loss function. It is a method for controlling forgetting by penalizing changes to the model parameters crucial for performance in tasks previously acquired. The regularization term is directly linked to the square of the difference between each parameter's current value and its value when the previous task was learned, adjusted according to the parameter's significance in the previous task. The regularization term corresponding to task $i$ is defined as:

\begin{equation}
\mathcal{L}_{\text{reg}, i} = \frac{\lambda}{2} \sum{k} M_{i, k} (w_k - w_{i, k}^{*})^2
\end{equation}
where $M_{i, k}$ is the Fisher Information Matrix (FIM) $k$ for task $i$. It plays a crucial role in the regularization loss term, as it determines the importance $M_{i, k}$ of parameter $k$ for task $i$. $w_k$ is the current value of parameter $k$, $w_{i, k}^{*}$ is the value of parameter $k$ when task $i$ was learned, and $\lambda$ is a hyper-parameter that regulates the magnitude of the penalty.

We compute the FIM for each model parameter to measure their importance for a given task. It is done by calculating the following for each parameter $k$:
\begin{equation}
M_{i, k} =  \left( \frac{\partial \log p(r_{t+1:T} | s_{1:t}, a_{1:t}, r_{1:t}, p_i)}{\partial w_k} \right)^2 
\end{equation}

The total loss function for the P2DT, including the regularization term, is then:

\begin{equation}
\mathcal{L} = \sum_{i=1}^{N} \mathcal{L}_i + \sum\limits_{i=1}^{N} \mathcal{L}_{\text{reg}, i}
\end{equation}
This loss function encourages the P2DT to accurately predict future rewards for each task, given past and current states, actions, and task prompts, while also penalizing changes to the model parameters that could harm performance on previously learned tasks. 

% \vspace{-1em}
\begin{algorithm}
\caption{Progressive Prompt Decision Transformer (P2DT)}
\label{alg:ProgressivePromptDecisionTransformer}
\begin{algorithmic}[1]
\Require Task trajectories $\mathcal{T}$
\Ensure Optimized general and expert attention blocks $G$ and $E$, and prompt tokens $p_i$ for each task $i$

\State Initialize general attention blocks $G$ and expert attention blocks $E$
\For{$i = 1, 2, \ldots, |T|$}
    \State Initialize prompt token $p_i$ for task $i$
    \For{each trajectory $t$ in task $T_i$}
        \State $e_t \gets \text{Embedding}(t)$
        \State $x_{g} \gets \text{G}(e_t)$
        \State $x_{e} \gets \text{E}(x_{g}, p_i)$
        \State $a \gets \text{Prediction}(x_{e})$
        \State $L \gets \text{ComputeLoss}(a, a_{target})$
        \State $G, E, p_i \gets \text{Backpropagate}(G, E, p_i, L)$
    \EndFor 
    \State Store $p_i$ and its corresponding state in $E$
\EndFor
\State For a new task $T_{new}$ at test time, use the stored $p_{new}$ and state in $E$ for prediction
\Return $G, E, p_i$
\end{algorithmic}
\end{algorithm}

\section{evaluation}
\label{sec:eval}
\subsection{Experiment Setup }
\label{ssec:exp}

\begin{table}[htb]
\centering
\caption{The D4RL score of our proposed algorithm P2DT under non-incremental (Non-inc.) and incremental (Inc.) settings.}
\label{tab:experiment_results}
\begin{threeparttable}
\begin{tabular}{
    @{\hspace{0pt}}l@{\hspace{3pt}} % Left column with space adjustment
    @{\hspace{3pt}}l@{\hspace{3pt}} % Method column with space adjustment
    S[table-format=2.1] % Ha column with adjusted number format
    S[table-format=2.1] % Ho column with adjusted number format
    S[table-format=2.1] % W column with adjusted number format
}
\toprule
\textbf{Settings} & \textbf{Method} & \textbf{Ha} & \textbf{Ho} & \textbf{W} \\ 
\midrule
\multirow{6}{*}{Non-inc.}
& DT & 42.6 & 67.6 & 74.0 \\ 
& CQL & 44.4 & 58.0 & 79.2 \\ 
& BEAR & 41.7 & 52.1 & 59.1 \\ 
& BRAC-v & 46.3 & 31.1 & 81.1\\
& AWR & 37.4 & 35.9 & 17.4 \\
& BC & 43.1 & 63.9 & 77.3 \\
& P2DT\tnote{*} & {42.7} & {56.3} & {79.8} \\
\addlinespace % Adds some space before the next line

\toprule
\textbf{Settings} & \textbf{Metrics} & \textbf{Ha-Ho-W} & \textbf{W-Ho-Ha} & \textbf{W-Ha-Ho} \\ 
\midrule
\multirow{3}{*}{Inc. (DT)}
& First & {1.2} & {-0.2} & {-0.2} \\ 
& Middle & {2.1} & {3.6} & {1.1} \\ 
& Last & {74.9} & {42.4} & {57.6} \\ 
\midrule
\multirow{3}{*}{Inc. (P2DT)}
& First & {36.8} & {66.2} & {73.6} \\ 
& Middle & {36.4} & {50.8} & {32.1} \\ 
& Last & {58.4} & {31.2} & {33.8} \\ 
\bottomrule
\end{tabular}
\begin{tablenotes}
\item[*] Ha: HalfCheetah, Ho: Hopper, W: Walker2D.
\item[*] `P2DT*' indicates the experiment setting is non-incremental.
\item[*] Prompt token length is 20. Learning steps are 20000. The numbers of GAB and EAB are 2 and 3, respectively.
\item[*] `First' represents the performance on the first dataset, `Middle' represents the middle datasets, and `Last' represents the last dataset.
\end{tablenotes}
\end{threeparttable}
\end{table}

To thoroughly assess the P2DT algorithm and compare it with the baselines, we conduct a series of experiments on three continuous control environments: HalfCheetah, Hopper, and Walker2D in the Gym MuJoCo environment \cite{todorov2012mujoco}. The same model learns these tasks in sequence to simulate a continual learning setting. We conduct experiments on the medium dataset from D4RL \cite{fu2020d4rl} for each of the three MuJoCo environments. 

During the learning phase, the model learns each task in sequence, starting with HalfCheetah, then moving to Hopper, and finally, Walker2D. For each task, the model is trained on the corresponding dataset from D4RL. We evaluate the model's performance on each task after learning all tasks.

\begin{figure}[htb]
    \begin{subfigure}{.45\linewidth}
        \centering
        \includegraphics[width=\linewidth]{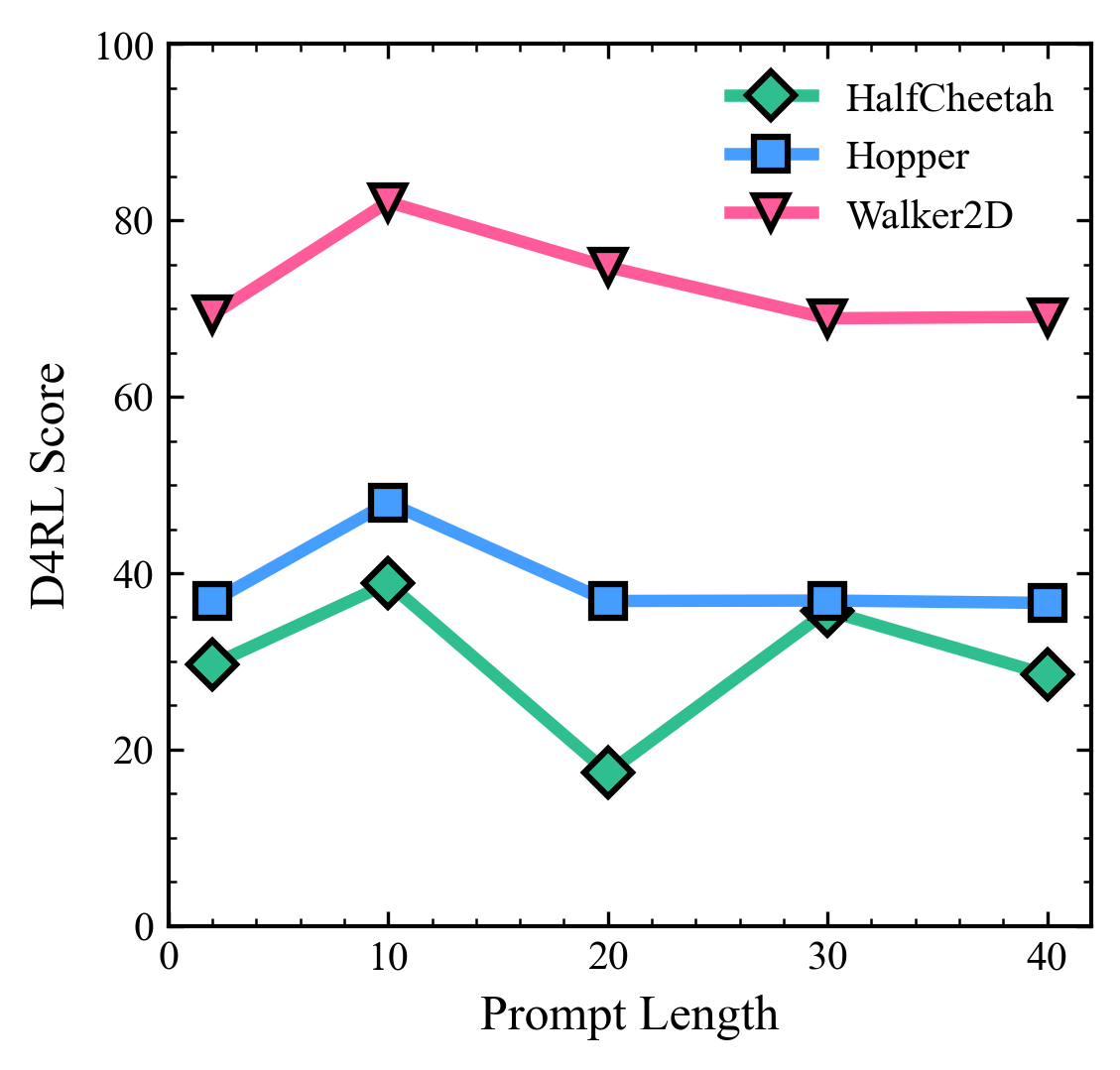}
        \caption{Analysis of D4RL score vs prompt length.}
        \label{fig:sub1}
    \end{subfigure}
    \hfill
    \begin{subfigure}{.45\linewidth}
        \centering      
        \includegraphics[width=\linewidth]{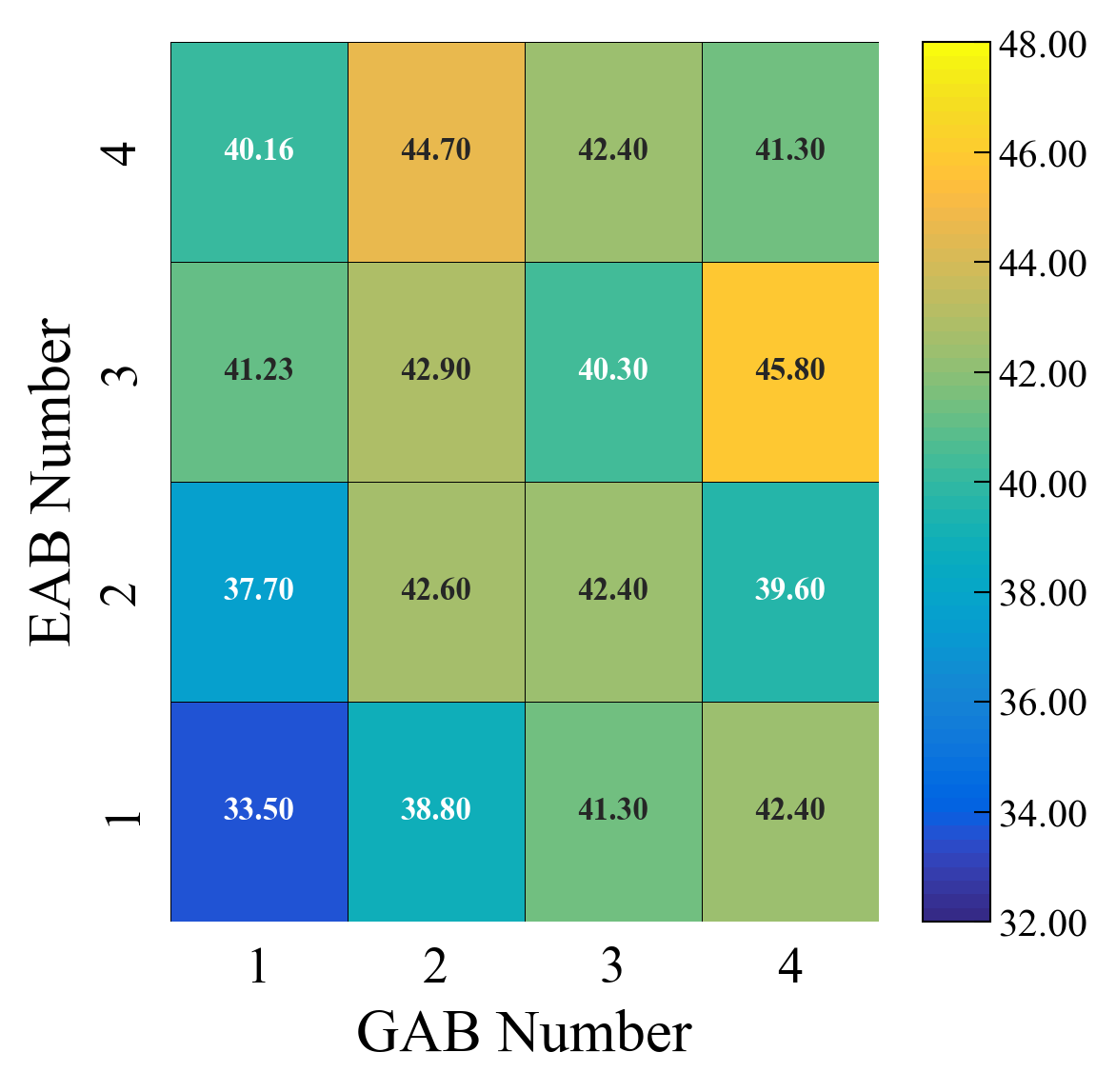}
        \caption{Grid search on the number of attention blocks.}
        \label{fig:sub2}
    \end{subfigure}

    \caption{The experimental results of our proposed algorithm P2DT in incremental setting.}
    \label{fig:res}
\end{figure}

% \vspace{-1.1em}

\subsection{Results Analysis}
\label{ssec:analysis}

As shown in Table \ref{tab:experiment_results}, we compared our proposed algorithm under non-incremental settings with several existing methods, including DT\cite{chen2021decision}, CQL\cite{kumar2020conservative}, BEAR\cite{kumar2019stabilizing}, BRAC\cite{wu2019behavior}, AWR\cite{peng2019advantage} and behavior cloning(BC). P2DT matches other methods in performance.
In the incremental learning setting, We tested three learning sequences, and while DT almost entirely forgets earlier tasks after completing the sequence, P2DT remarkably preserves prior knowledge. This underscores P2DT's superior capability in retaining information in a task-incremental learning framework.

We calculated the average scores of these algorithms on the medium dataset. Due to significant fluctuations in the scores of these algorithms, we selected the highest score achieved during their respective runs. As illustrated in table \ref{tab:experiment_results}, P2DT achieves performance surpassing the majority of algorithms in the task incremental learning scenario, with only a minor performance loss compared to the original Decision Transformer. These features allow P2DT to handle the complexities associated with personalized agents more effectively than other methods. Figure \ref{fig:sub1} aims to highlight that, as prompt length increases, accuracy reaches a saturation point. Figure \ref{fig:sub2} shows the influence of the number of GABs and EABs on the final results.

\begin{table}[!ht]
\caption{Parameter analysis of P2DT vs Decision Transformer}
\label{tab:parameter size}
\resizebox{\columnwidth}{!}{
\begin{tabular}{cccc}
\toprule
\textbf{Method} & \textbf{Length} & \textbf{Param.} & \textbf{Size (KB)} \\

\midrule
HalfCheetah / Walker2D(DT) & Total & 27.73M & 29215 \\
\midrule
Hopper(DT) & Total & 27.72M & 29204 \\
\midrule
\multirow{3}{*}{Prompts of P2DT} & 5 & 1.92k & 7.5 \\
& 10 & 3.84k & 15 \\
& 20 & 7.68k & 30 \\

\bottomrule
\end{tabular}
}
\end{table}

\subsection{Consumption Analysis}
\label{ssec:consumption analysis}

A consumption analysis of the P2DT model was conducted, focusing on the parameter count (Table \ref{tab:parameter size}). We compute our cost assuming the number of EAB is 3. Our approach can scale to many tasks due to the careful management of parameter expansion. Each new task only requires the addition of a new token, eliminating the need for retraining the model. Compared to the original decision transformer, this efficiency results in a practically negligible overhead. The computational complexity of P2DT is primarily determined by the operations in the Transformer decode. 

\section{Conclusion}
\label{sec:typestyle}
In conclusion, our work presented the Progressive Prompt Decision Transformer, a novel solution to the catastrophic forgetting problem in multi-task learning environments. Our algorithm successfully balances generalization and specialization by deploying General Attention Blocks and Expert Attention Blocks. Empirical results demonstrated superior performance and cost-effectiveness compared to existing methods.

\section{ACKNOWLEDGEMENT}
\label{sec:ack}

Supported by the Key Research and Development Program of Guangdong Province under grant No.2021B0101400003.

\vfill\pagebreak
\label{sec:refs}

\bibliographystyle{IEEE}
\bibliography{refs}

\end{document}